%% file: end-to-end-learning-for-self-driving-cars.tex
\documentclass{article} 
\usepackage[hidelinks]{hyperref}
\usepackage{url}
\usepackage{graphicx}           
\usepackage{nvmacros}
\usepackage{glossaries}
\usepackage{amsmath}
\makeglossaries
\usepackage{nips15submit_e,times}
\nipsfinalcopy
\title{End to End Learning for Self-Driving Cars}
\author{
    Mariusz Bojarski \\
    NVIDIA Corporation \\
    Holmdel, NJ 07735 \\
  \And
    Davide Del Testa \\
    NVIDIA Corporation \\
    Holmdel, NJ 07735 \\
  \And 
    Daniel Dworakowski \\
    NVIDIA Corporation \\
    Holmdel, NJ 07735 \\
  \And
     Bernhard Firner \\
    NVIDIA Corporation \\
    Holmdel, NJ 07735 \\
  \And
    Beat Flepp \\
    NVIDIA Corporation \\
    Holmdel, NJ 07735 \\
  \And 
    Prasoon Goyal \\
    NVIDIA Corporation \\
    Holmdel, NJ 07735 \\
  \And
    Lawrence D. Jackel \\
    NVIDIA Corporation \\
    Holmdel, NJ 07735 \\
  \And 
    Mathew Monfort \\
    NVIDIA Corporation \\
    Holmdel, NJ 07735 \\
  \And
    Urs Muller \\
    NVIDIA Corporation \\
    Holmdel, NJ 07735 \\
  \And
    Jiakai Zhang \\
    NVIDIA Corporation \\
    Holmdel, NJ 07735 \\
  \And
    Xin Zhang \\
    NVIDIA Corporation \\
    Holmdel, NJ 07735 \\
  \And
    Jake Zhao \\
    NVIDIA Corporation \\
    Holmdel, NJ 07735 \\
  \And
    Karol Zieba \\
    NVIDIA Corporation \\
    Holmdel, NJ 07735 \\
}
\input acronyms
\begin{document}
\maketitle
%
\input body
%
\small
\bibliography{neural-nets}
\bibliographystyle{unsrturl}
%
\label{lastpage}
\end{document}

%% file: acronyms.tex
\newacronym{cnn}{CNN}{convolutional neural network}
\newacronym{ilsvrc}{ILSVRC}{Large Scale Visual Recognition Challenge}
\newacronym{cpu}{GPU}{Graphical Processing Unit}
\newacronym{darpa}{DARPA}{Defense Advanced Research Projects Agency}
\newacronym{dave}{DAVE}{\gls{darpa} Autonomous Vehicle}
\newacronym{rc}{RC}{radio control}
\newacronym{alvinn}{ALVINN}{Autonomous Land Vehicle in a Neural Network}
\newacronym{lagr}{LAGR}{Learning Applied to Ground Robots}
\newacronym{gpu}{GPU}{graphics processing unit}
\newacronym{can}{CAN}{Controller Area Network}
\newacronym{fps}{FPS}{frames per second}

%% file: body.tex
%
%
%
%
%
%
%

\begin{abstract}
We trained a \gls{cnn} to map raw pixels from a single front-facing
camera directly to steering commands. This end-to-end approach proved
surprisingly powerful. With minimum training data from humans the
system learns to drive in traffic on local roads with or without lane
markings and on highways. It also operates in areas with unclear
visual guidance such as in parking lots and on unpaved roads.

The system automatically learns internal representations of the
necessary processing steps such as detecting useful road features with
only the human steering angle as the training signal. We never
explicitly trained it to detect, for example, the outline of roads.

Compared to explicit decomposition of the problem, such as lane
marking detection, path planning, and control, our end-to-end system
optimizes all processing steps simultaneously.  We argue that this
will eventually lead to better performance and smaller systems. Better
performance will result because the internal components self-optimize
to maximize overall system performance, instead of optimizing
human-selected intermediate criteria, \eg, lane detection.  Such
criteria understandably are selected for ease of human interpretation
which doesn't automatically guarantee maximum system performance.
Smaller networks are possible because the system learns to solve the
problem with the minimal number of processing steps.

We used an NVIDIA DevBox and Torch~7 for training and an NVIDIA
\drivepx{} self-driving car computer also running Torch~7 for
determining where to drive. The system operates at 30 \gls{fps}.
\end{abstract}

\clearpage
\section{Introduction}
\glspl{cnn} \cite{lecun-89e} have revolutionized pattern recognition
\cite{krizhevsky-2012}. Prior to the widespread adoption of
\glspl{cnn}, most pattern recognition tasks were performed using an
initial stage of hand-crafted feature extraction followed by a
classifier. The breakthrough of \glspl{cnn} is that features are
learned automatically from training examples. The \gls{cnn} approach
is especially powerful in image recognition tasks because the
convolution operation captures the 2D nature of images. Also, by using
the convolution kernels to scan an entire image, relatively few
parameters need to be learned compared to the total number of
operations.

While \glspl{cnn} with learned features have been in commercial use
for over twenty years \cite{jackel-1995}, their adoption has exploded
in the last few years because of two recent developments.  First,
large, labeled data sets such as the \gls{ilsvrc} \cite{ilsvrc} have
become available for training and validation. Second, \gls{cnn}
learning algorithms have been implemented on the massively parallel
\glspl{gpu} which tremendously accelerate learning and inference.

In this paper, we describe a \gls{cnn} that goes beyond pattern
recognition. It learns the entire processing pipeline needed to steer
an automobile.  The groundwork for this project was done over 10 years
ago in a \gls{darpa} seedling project known as \gls{dave}
\cite{dave-report-04} in which a sub-scale \gls{rc} car drove through
a junk-filled alley way. \gls{dave} was trained on hours of human
driving in similar, but not identical environments. The training data
included video from two cameras coupled with left and right steering
commands from a human operator.

In many ways, DAVE-2 was inspired by the pioneering work of Pomerleau
\cite{pomerleau-1989} who in 1989 built the \gls{alvinn} system. It
demonstrated that an end-to-end trained neural network can indeed
steer a car on public roads. Our work differs in that 25 years of
advances let us apply far more data and computational power to the
task. In addition, our experience with \glspl{cnn} lets us make use of
this powerful technology. (\gls{alvinn} used a fully-connected network
which is tiny by today's standard.)

While \gls{dave} demonstrated the potential of end-to-end learning,
and indeed was used to justify starting the \gls{darpa} \gls{lagr}
program \cite{wikipedia-lagr}, \gls{dave}'s performance was not
sufficiently reliable to provide a full alternative to more modular
approaches to off-road driving. \gls{dave}'s mean distance between
crashes was about 20~meters in complex environments.

Nine months ago, a new effort was started at NVIDIA that sought to
build on DAVE and create a robust system for driving on public
roads. The primary motivation for this work is to avoid the need to
recognize specific human-designated features, such as lane markings,
guard rails, or other cars, and to avoid having to create a collection
of ``if, then, else'' rules, based on observation of these features.
This paper describes preliminary results of this new effort.

\section{Overview of the DAVE-2 System}
\label{sec-overview}
\begin{figure}[htb]
  \hfil
  \includegraphics[scale=0.85]{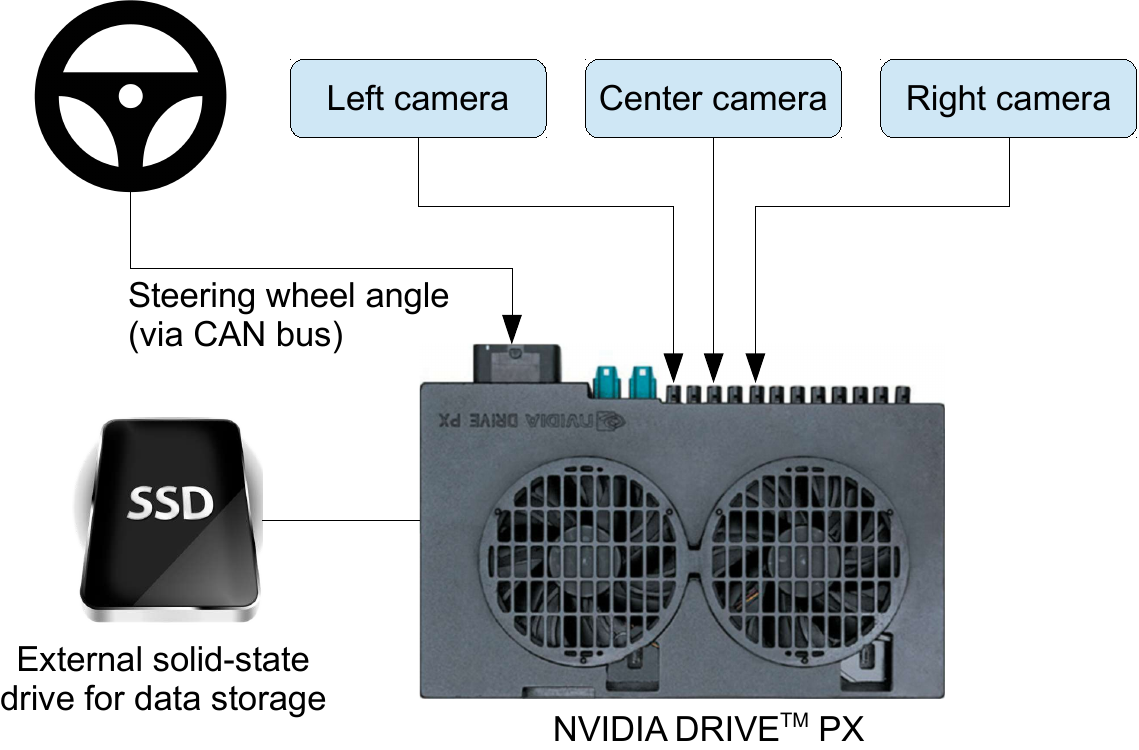}
  \caption{High-level view of the data collection system.}
  \label{fig-data-collection-system}
\end{figure}

Figure~\ref{fig-data-collection-system} shows a simplified block
diagram of the collection system for training data for \mbox{DAVE-2}.
Three cameras are mounted behind the windshield of the
data-acquisition car. Time-stamped video from the cameras is captured
simultaneously with the steering angle applied by the human
driver. This steering command is obtained by tapping into the
vehicle's \gls{can} bus. In order to make our system independent of
the car geometry, we represent the steering command as $^1/_r$ where
$r$ is the turning radius in meters. We use $^1/_r$ instead of $r$ to
prevent a singularity when driving straight (the turning radius for
driving straight is infinity).  $^1/_r$ smoothly transitions through
zero from left turns (negative values) to right turns (positive
values).

Training data contains single images sampled from the video, paired
with the corresponding steering command ($^1/_r$). Training with data
from only the human driver is not sufficient. The network must learn
how to recover from mistakes. Otherwise the car will slowly drift off
the road. The training data is therefore augmented with additional
images that show the car in different shifts from the center of the
lane and rotations from the direction of the road.

Images for two specific off-center shifts can be obtained from the
left and the right camera. Additional shifts between the cameras and
all rotations are simulated by viewpoint transformation of the image
from the nearest camera. Precise viewpoint transformation requires 3D
scene knowledge which we don't have. We therefore approximate the
transformation by assuming all points below the horizon are on flat
ground and all points above the horizon are infinitely far away. This
works fine for flat terrain but it introduces distortions for objects
that stick above the ground, such as cars, poles, trees, and
buildings. Fortunately these distortions don't pose a big problem for
network training. The steering label for transformed images is
adjusted to one that would steer the vehicle back to the desired
location and orientation in two seconds.

\begin{figure}[htb]
  \hfil
  \includegraphics[scale=0.85]{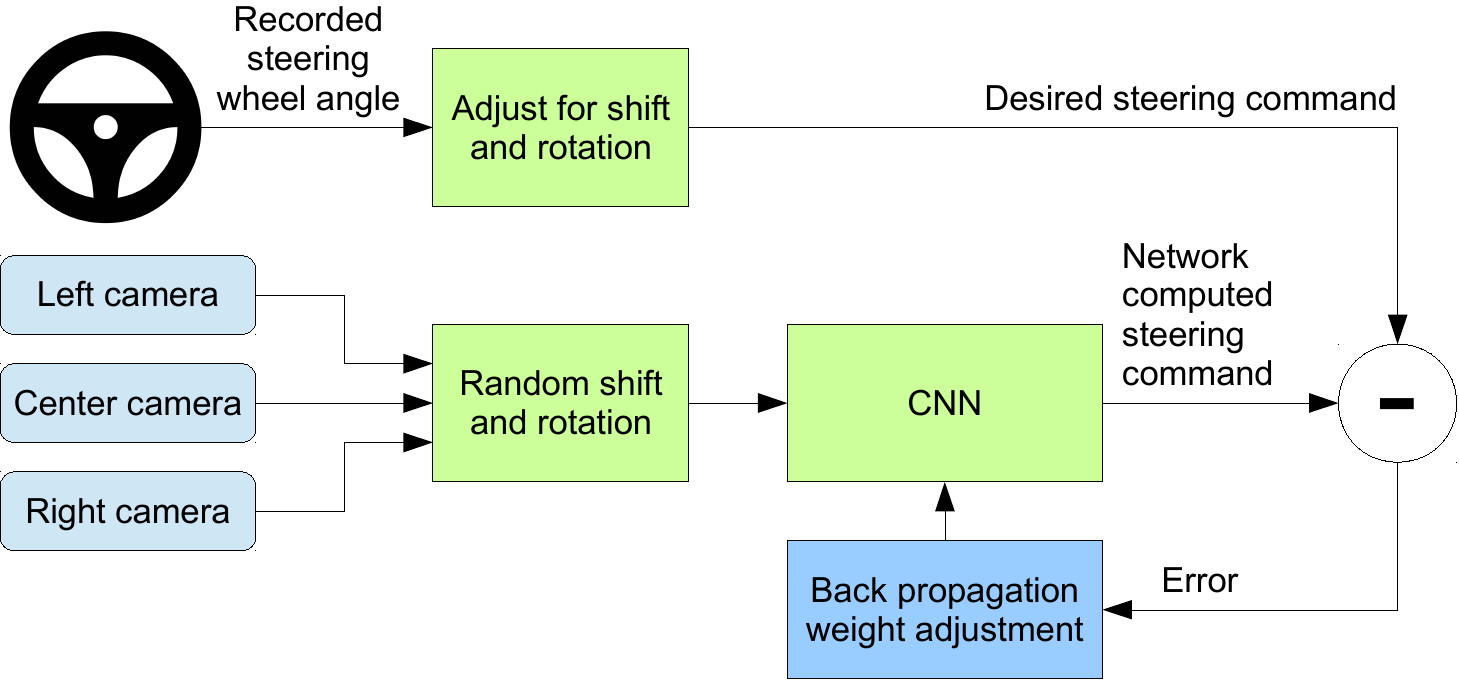}
  \caption{Training the neural network.}
  \label{fig-training}
\end{figure}

A block diagram of our training system is shown in
Figure~\ref{fig-training}.  Images are fed into a \gls{cnn} which then
computes a proposed steering command.  The proposed command is
compared to the desired command for that image and the weights of the
\gls{cnn} are adjusted to bring the \gls{cnn} output closer to the
desired output. The weight adjustment is accomplished using back
propagation as implemented in the Torch~7 machine learning package.

\begin{figure}[htb]
  \hfil
  \includegraphics[scale=0.85]{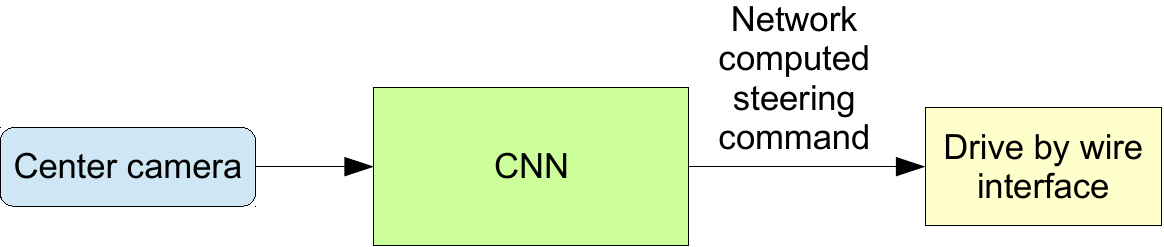}
  \caption{The trained network is used to generate steering commands
    from a single front-facing center camera.}
  \label{fig-inference}
\end{figure}

Once trained, the network can generate steering from the video images
of a single center camera. This configuration is shown in
Figure~\ref{fig-inference}.

\section{Data Collection}
Training data was collected by driving on a wide variety of roads and
in a diverse set of lighting and weather conditions. Most road data
was collected in central New Jersey, although highway data was also
collected from Illinois, Michigan, Pennsylvania, and New York. Other
road types include two-lane roads (with and without lane markings),
residential roads with parked cars, tunnels, and unpaved roads. Data
was collected in clear, cloudy, foggy, snowy, and rainy weather, both
day and night. In some instances, the sun was low in the sky,
resulting in glare reflecting from the road surface and scattering
from the windshield.

Data was acquired using either our drive-by-wire test vehicle, which
is a 2016 Lincoln MKZ, or using a 2013 Ford Focus with cameras placed
in similar positions to those in the Lincoln. The system has no
dependencies on any particular vehicle make or model. Drivers were
encouraged to maintain full attentiveness, but otherwise drive as they
usually do. As of March 28, 2016, about 72 hours of driving data was
collected.

\section{Network Architecture}
\begin{figure}[htb]
  \hfil
  \includegraphics[scale=0.55]{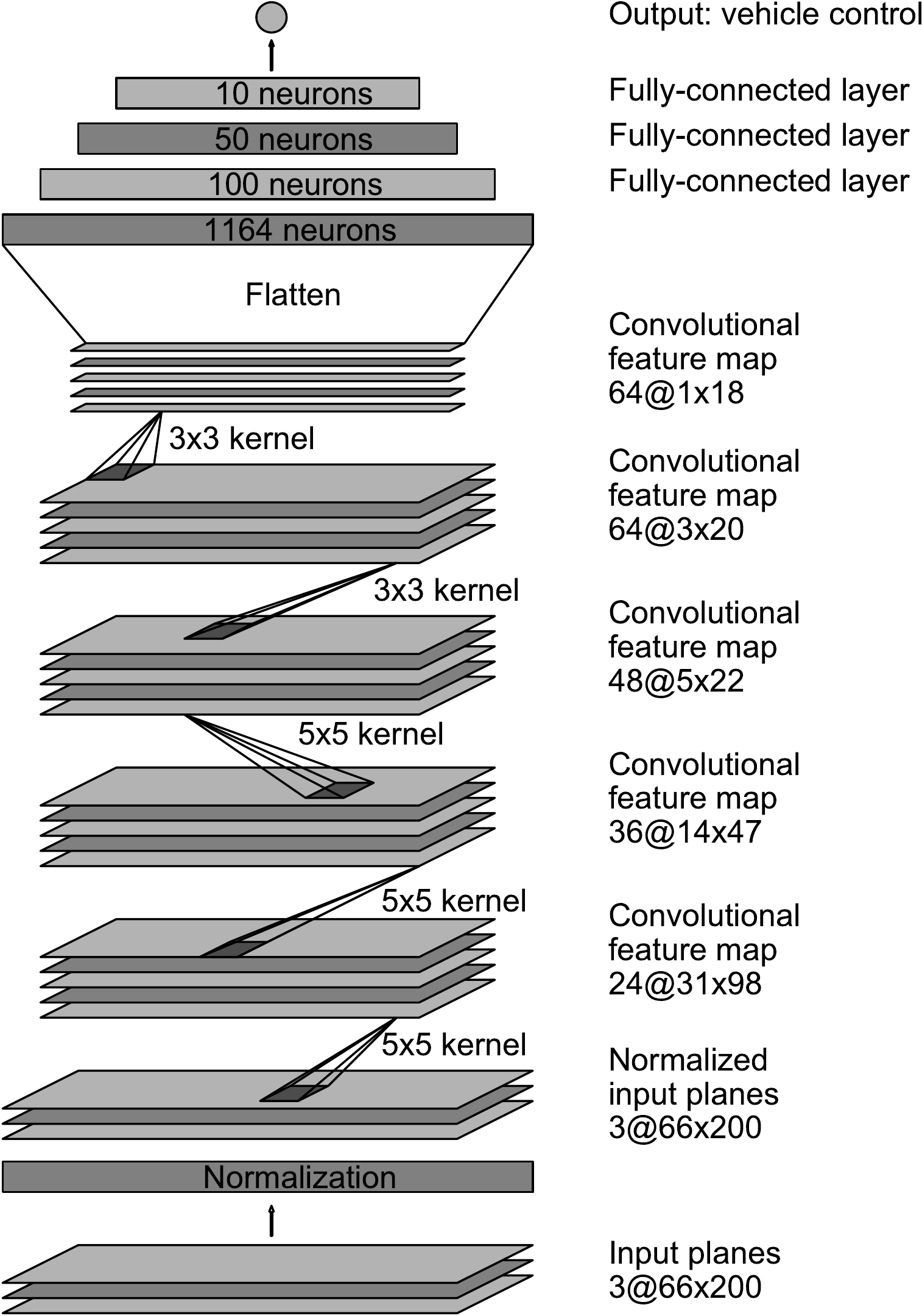}
  \caption{\gls{cnn} architecture. The network has about 27 million
    connections and 250 thousand parameters.}
  \label{fig-cnn-architecture}
\end{figure}

We train the weights of our network to minimize the mean squared error
between the steering command output by the network and the command of
either the human driver, or the adjusted steering command for
off-center and rotated images (see Section~\ref{sec-augmentation}).
Our network architecture is shown in
Figure~\ref{fig-cnn-architecture}.  The network consists of 9 layers,
including a normalization layer, 5 convolutional layers and 3 fully
connected layers. The input image is split into YUV planes and passed
to the network.

The first layer of the network performs image normalization. The
normalizer is hard-coded and is not adjusted in the learning process.
Performing normalization in the network allows the normalization
scheme to be altered with the network architecture and to be
accelerated via \gls{gpu} processing.

The convolutional layers were designed to perform feature extraction
and were chosen empirically through a series of experiments that
varied layer configurations. We use strided convolutions in the first
three convolutional layers with a 2\x2 stride and a 5\x5 kernel and a
non-strided convolution with a 3\x3 kernel size in the last two
convolutional layers.

We follow the five convolutional layers with three fully connected
layers leading to an output control value which is the inverse turning
radius. The fully connected layers are designed to function as a
controller for steering, but we note that by training the system
end-to-end, it is not possible to make a clean break between which
parts of the network function primarily as feature extractor and which
serve as controller.

\section{Training Details}

\subsection{Data Selection}
The first step to training a neural network is selecting the frames to
use. Our collected data is labeled with road type, weather condition,
and the driver's activity (staying in a lane, switching lanes,
turning, and so forth). To train a \gls{cnn} to do lane following we
only select data where the driver was staying in a lane and discard
the rest. We then sample that video at 10~\gls{fps}. A higher
sampling rate would result in including images that are highly similar
and thus not provide much useful information.

To remove a bias towards driving straight the training data includes a
higher proportion of frames that represent road curves.

\subsection{Augmentation}
\label{sec-augmentation}
After selecting the final set of frames we augment the data by adding
artificial shifts and rotations to teach the network how to recover
from a poor position or orientation. The magnitude of these
perturbations is chosen randomly from a normal distribution. The
distribution has zero mean, and the standard deviation is twice the
standard deviation that we measured with human drivers.  Artificially
augmenting the data does add undesirable artifacts as the magnitude
increases (see Section~\ref{sec-overview}).

\section{Simulation}
\label{sec-simulation}
\begin{figure}[htb]
  \hfil
  \includegraphics[scale=0.85]{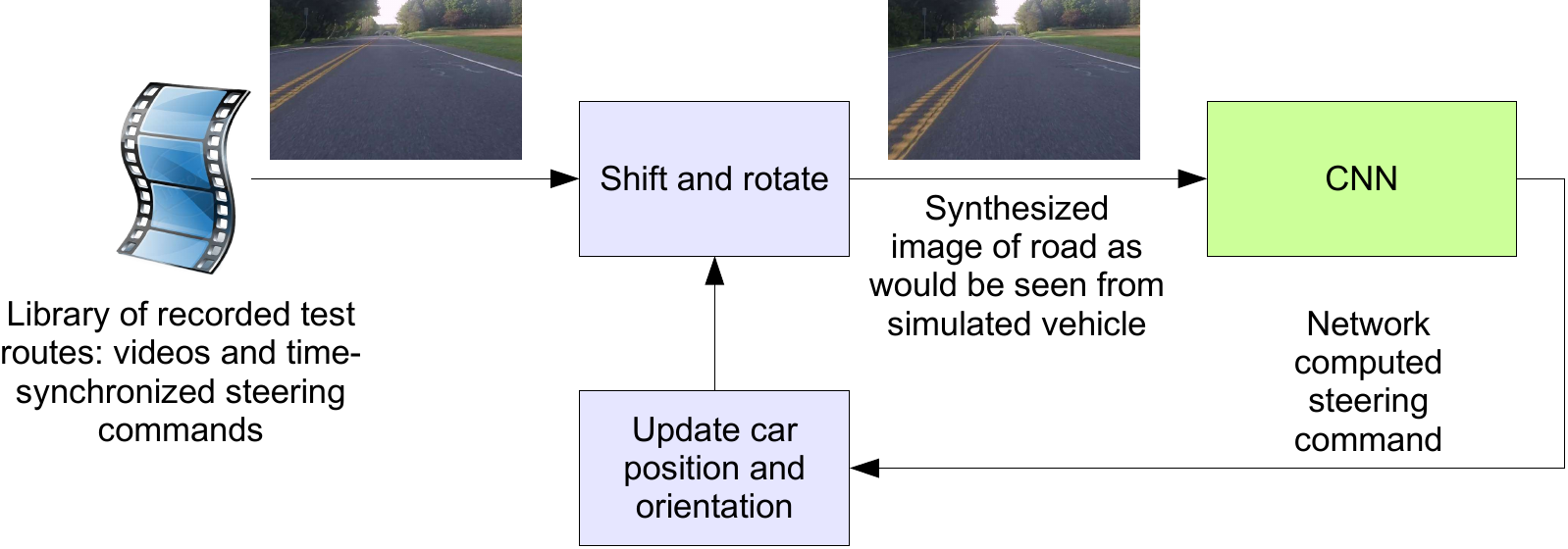}
  \caption{Block-diagram of the drive simulator.}
  \label{fig-simulator}
\end{figure}

Before road-testing a trained \gls{cnn}, we first evaluate the network’s
performance in simulation. A simplified block diagram of the
simulation system is shown in Figure~\ref{fig-simulator}. 

The simulator takes pre-recorded videos from a forward-facing on-board
camera on a human-driven data-collection vehicle and generates images
that approximate what would appear if the \gls{cnn} were, instead,
steering the vehicle. These test videos are time-synchronized with
recorded steering commands generated by the human driver.

Since human drivers might not be driving in the center of the lane all
the time, we manually calibrate the lane center associated with each
frame in the video used by the simulator. We call this position the
``ground truth''. 

The simulator transforms the original images to account for departures
from the ground truth. Note that this transformation also includes any
discrepancy between the human driven path and the ground truth. The
transformation is accomplished by the same methods described in
Section~\ref{sec-overview}.

The simulator accesses the recorded test video along with the
synchronized steering commands that occurred when the video was
captured. The simulator sends the first frame of the chosen test
video, adjusted for any departures from the ground truth, to the input
of the trained \gls{cnn}. The \gls{cnn} then returns a steering
command for that frame.  The \gls{cnn} steering commands as well as
the recorded human-driver commands are fed into the dynamic model
\cite{wang-2001} of the vehicle to update the position and orientation
of the simulated vehicle.

The simulator then modifies the next frame in the test video so that
the image appears as if the vehicle were at the position that resulted
by following steering commands from the \gls{cnn}. This new image is then
fed to the \gls{cnn} and the process repeats.

The simulator records the off-center distance (distance from the car
to the lane center), the yaw, and the distance traveled by the virtual
car. When the off-center distance exceeds one meter, a virtual “human
intervention” is triggered, and the virtual vehicle position and
orientation is reset to match the ground truth of the corresponding
frame of the original test video.

\begin{figure}[htb]
  \hfil
  \includegraphics[width=1.0\textwidth,height=.4717\textwidth]{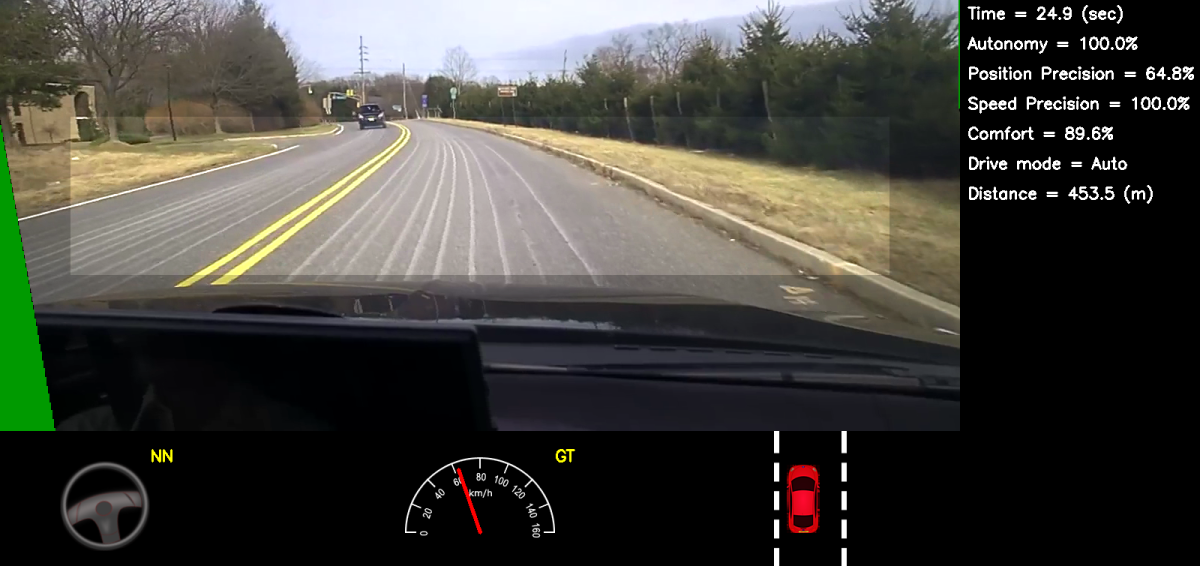}
  \caption{Screen shot of the simulator in interactive mode. See
    Section~\ref{sec-sim-tests} for explanation of the performance
    metrics. The green area on the left is unknown because of the
    viewpoint transformation. The highlighted wide rectangle below the
    horizon is the area which is sent to the \gls{cnn}.}
\end{figure}

\section{Evaluation}
\label{sec-evaluation}
Evaluating our networks is done in two steps, first in simulation, and
then in on-road tests.

In simulation we have the networks provide steering commands in our
simulator to an ensemble of prerecorded test routes that correspond to
about a total of three hours and 100~miles of driving in Monmouth
County, NJ.  The test data was taken in diverse lighting and weather
conditions and includes highways, local roads, and residential
streets.

\subsection{Simulation Tests}
\label{sec-sim-tests}
We estimate what percentage of the time the network could drive the
car (autonomy).  The metric is determined by counting simulated “human
interventions” (see Section~\ref{sec-simulation}). These interventions
occur when the simulated vehicle departs from the center line by more
than one meter. We assume that in real life an actual intervention
would require a total of six seconds: this is the time required for a
human to retake control of the vehicle, re-center it, and then restart
the self-steering mode. We calculate the percentage autonomy by
counting the number of interventions, multiplying by 6 seconds,
dividing by the elapsed time of the simulated test, and then
subtracting the result from 1:
\begin{equation}
  \label{eq-autonomy}
  \text{autonomy}=(1-\frac{(\text{number of interventions})
    \cdot6\text{ seconds}}
    {\text{elapsed time [seconds]}})\cdot100
\end{equation}
Thus, if we had 10 interventions in 600 seconds, we would have an
autonomy value of
\begin{equation*}
  (1-\frac{10\cdot6}{600})\cdot100=90\%
\end{equation*}

\subsection{On-road Tests}
After a trained network has demonstrated good performance in the
simulator, the network is loaded on the \drivepx{} in our test car and
taken out for a road test. For these tests we measure performance as
the fraction of time during which the car performs autonomous
steering. This time excludes lane changes and turns from one road to
another.  For a typical drive in Monmouth County NJ from our office in
Holmdel to Atlantic Highlands, we are autonomous approximately 98\% of
the time. We also drove 10~miles on the Garden State Parkway (a
multi-lane divided highway with on and off ramps) with zero
intercepts.

A video of our test car driving in diverse conditions can be seen in
\cite{dave2-video}.

\begin{figure}[p]
  \hfil
  \includegraphics{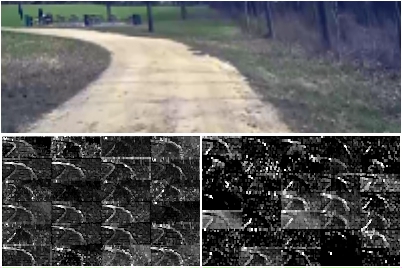}
  \caption{How the \gls{cnn} ``sees'' an unpaved road. Top: subset of
    the camera image sent to the \gls{cnn}. Bottom left: Activation of
    the first layer feature maps. Bottom right: Activation of the
    second layer feature maps. This demonstrates that the \gls{cnn}
    learned to detect useful road features on its own, \ie, with only
    the human steering angle as training signal. We never explicitly
    trained it to detect the outlines of roads.}
  \label{fig-feature-maps-road}
\end{figure}

\begin{figure}[p]
  \hfil
  \includegraphics{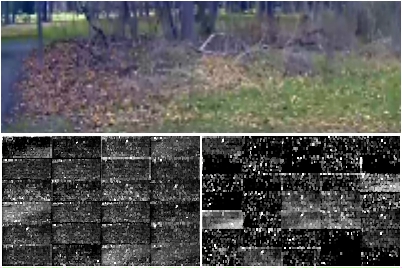}
  \caption{Example image with no road. The activations of the first
    two feature maps appear to contain mostly noise, \ie, the
    \gls{cnn} doesn't recognize any useful features in this image.}
  \label{fig-feature-maps-woods}
\end{figure}

\subsection{Visualization of Internal \gls{cnn} State}
Figures~\ref{fig-feature-maps-road} and ~\ref{fig-feature-maps-woods}
show the activations of the first two feature map layers for two
different example inputs, an unpaved road and a forest. In case of the
unpaved road, the feature map activations clearly show the outline of
the road while in case of the forest the feature maps contain mostly
noise, \ie, the \gls{cnn} finds no useful information in this image.

This demonstrates that the \gls{cnn} learned to detect useful road
features on its own, \ie, with only the human steering angle as
training signal. We never explicitly trained it to detect the outlines
of roads, for example.

\clearpage
\section{Conclusions}
We have empirically demonstrated that \glspl{cnn} are able to learn
the entire task of lane and road following without manual
decomposition into road or lane marking detection, semantic
abstraction, path planning, and control. A small amount of training
data from less than a hundred hours of driving was sufficient to train
the car to operate in diverse conditions, on highways, local and
residential roads in sunny, cloudy, and rainy conditions. The
\gls{cnn} is able to learn meaningful road features from a very sparse
training signal (steering alone).

The system learns for example to detect the outline of a road without
the need of explicit labels during training.

More work is needed to improve the robustness of the network, to find
methods to verify the robustness, and to improve visualization of the
network-internal processing steps.

%% file: end-to-end-learning-for-self-driving-cars.bbl
\begin{thebibliography}{1}

\bibitem{lecun-89e}
Y.~LeCun, B.~Boser, J.~S. Denker, D.~Henderson, R.~E. Howard, W.~Hubbard, and
  L.~D. Jackel.
\newblock Backpropagation applied to handwritten zip code recognition.
\newblock {\em Neural Computation}, 1(4):541--551, Winter 1989.
\newblock URL: \url{http://yann.lecun.org/exdb/publis/pdf/lecun-89e.pdf}.

\bibitem{krizhevsky-2012}
Alex Krizhevsky, Ilya Sutskever, and Geoffrey~E. Hinton.
\newblock Imagenet classification with deep convolutional neural networks.
\newblock In F.~Pereira, C.~J.~C. Burges, L.~Bottou, and K.~Q. Weinberger,
  editors, {\em Advances in Neural Information Processing Systems 25}, pages
  1097--1105. Curran Associates, Inc., 2012.
\newblock URL:
  \url{http://papers.nips.cc/paper/4824-imagenet-classification-with-deep-convolutional-neural-networks.pdf}.

\bibitem{jackel-1995}
L.~D. Jackel, D.~Sharman, Stenard~C. E., Strom~B. I., , and D~Zuckert.
\newblock Optical character recognition for self-service banking.
\newblock {\em AT\&T Technical Journal}, 74(1):16--24, 1995.

\bibitem{ilsvrc}
Large scale visual recognition challenge ({ILSVRC}).
\newblock URL: \url{http://www.image-net.org/challenges/LSVRC/}.

\bibitem{dave-report-04}
{Net-Scale Technologies, Inc.}
\newblock Autonomous off-road vehicle control using end-to-end learning, July
  2004.
\newblock Final technical report.
\newblock URL: \url{http://net-scale.com/doc/net-scale-dave-report.pdf}.

\bibitem{pomerleau-1989}
Dean~A. Pomerleau.
\newblock {ALVINN}, an autonomous land vehicle in a neural network.
\newblock Technical report, Carnegie Mellon University, 1989.
\newblock URL:
  \url{http://repository.cmu.edu/cgi/viewcontent.cgi?article=2874&context=compsci}.

\bibitem{wikipedia-lagr}
Wikipedia.org.
\newblock {DARPA} {LAGR} program.
\newblock \url{http://en.wikipedia.org/wiki/DARPA_LAGR_Program}.

\bibitem{wang-2001}
Danwei Wang and Feng Qi.
\newblock Trajectory planning for a four-wheel-steering vehicle.
\newblock In {\em Proceedings of the 2001 IEEE International Conference on
  Robotics \& Automation}, May 21--26 2001.
\newblock URL:
  \url{http://www.ntu.edu.sg/home/edwwang/confpapers/wdwicar01.pdf}.

\bibitem{dave2-video}
{DAVE} 2 driving a lincoln.
\newblock URL:
  \url{https://drive.google.com/open?id=0B9raQzOpizn1TkRIa241ZnBEcjQ}.

\end{thebibliography}
